\documentclass[10pt]{article}

\usepackage[preprint]{tmlr}

\usepackage{amsmath,amsfonts,bm}

\def\eqref#1{equation~\ref{#1}}
\def\1{\bm{1}}

\DeclareMathAlphabet{\mathsfit}{\encodingdefault}{\sfdefault}{m}{sl}
\SetMathAlphabet{\mathsfit}{bold}{\encodingdefault}{\sfdefault}{bx}{n}

\newcommand{\boldx}{{\boldsymbol{x}}}

\usepackage{hyperref}
\usepackage{url}
\usepackage{graphicx, color}

\usepackage[noorphans]{quoting}
\usepackage[square,sort,comma,numbers]{natbib}

\usepackage{amsmath}
\usepackage{amssymb}
\usepackage{mathtools}
\usepackage{amsthm}
\usepackage{bbm}
\usepackage{booktabs}
\usepackage{subcaption}
\usepackage{wrapfig}

\usepackage{caption}

\usepackage[most]{tcolorbox}

\usepackage[ruled, algo2e, vlined]{algorithm2e}
\SetKwInput{KwInput}{Input}
\SetKwInput{KwOutput}{Output}

\theoremstyle{plain}

\theoremstyle{definition}

\theoremstyle{remark}

\allowdisplaybreaks

\usepackage{listings}
\usepackage{xcolor}
\definecolor{codegreen}{rgb}{0,0.6,0}
\definecolor{codegray}{rgb}{0.5,0.5,0.5}
\definecolor{codepurple}{rgb}{0.58,0,0.82}
\definecolor{backcolour}{rgb}{1,1,1}

\lstdefinestyle{mystyle}{
    backgroundcolor=\color{backcolour},   
    commentstyle=\color{codegreen},
    keywordstyle=\color{magenta},
    numberstyle=\tiny\color{codegray},
    stringstyle=\color{codepurple},
    basicstyle=\ttfamily\footnotesize, % font size
    breakatwhitespace=false,         
    breaklines=true,                 
    captionpos=b,                    
    keepspaces=true,                 
    numbers=left,                    
    numbersep=5pt,                  
    showspaces=false,                
    showstringspaces=false,
    showtabs=false,                  
    tabsize=4,
    frame=lines      % lines at top and bottom
}

\title{Even GPT-5.2 Can't Count to Five:\\The Case for Zero-Error Horizons in Trustworthy LLMs}

\author{\name Ryoma Sato \email rsato@nii.ac.jp \\
  \addr National Institute of Informatics
}

\begin{document}

\maketitle

\begin{abstract}
We propose Zero-Error Horizon (ZEH) for trustworthy LLMs, which represents the maximum range that a model can solve without any errors. While ZEH itself is simple, we demonstrate that evaluating the ZEH of state-of-the-art LLMs yields abundant insights. For example, by evaluating the ZEH of GPT-5.2, we found that GPT-5.2 cannot even compute the parity of a short string like 11000, and GPT-5.2 cannot determine whether the parentheses in ((((()))))) are balanced. This is surprising given the excellent capabilities of GPT-5.2. The fact that LLMs make mistakes on such simple problems serves as an important lesson when applying LLMs to safety-critical domains. By applying ZEH to Qwen2.5 and conducting detailed analysis, we found that while ZEH correlates with accuracy, the detailed behaviors differ, and ZEH provides clues about the emergence of algorithmic capabilities. Finally, while computing ZEH incurs significant computational cost, we discuss how to mitigate this cost by achieving up to one order of magnitude speedup using tree structures and online softmax.
\end{abstract}

\section{Introduction}

Large Language Models (LLMs) are being applied to an increasingly wide range of domains. In particular, their use in safety-critical domains such as healthcare, law, finance, and science is being actively explored \cite{aljohani2025trustworthyllmhealthcare,dehghani2025large,mahdavi2025llmfinancial,tang2025financereasoning,yamada2025aiscientist,wei2025agenticscience}.

\begin{figure}[t]
  \centering
  \includegraphics[width=0.85\textwidth]{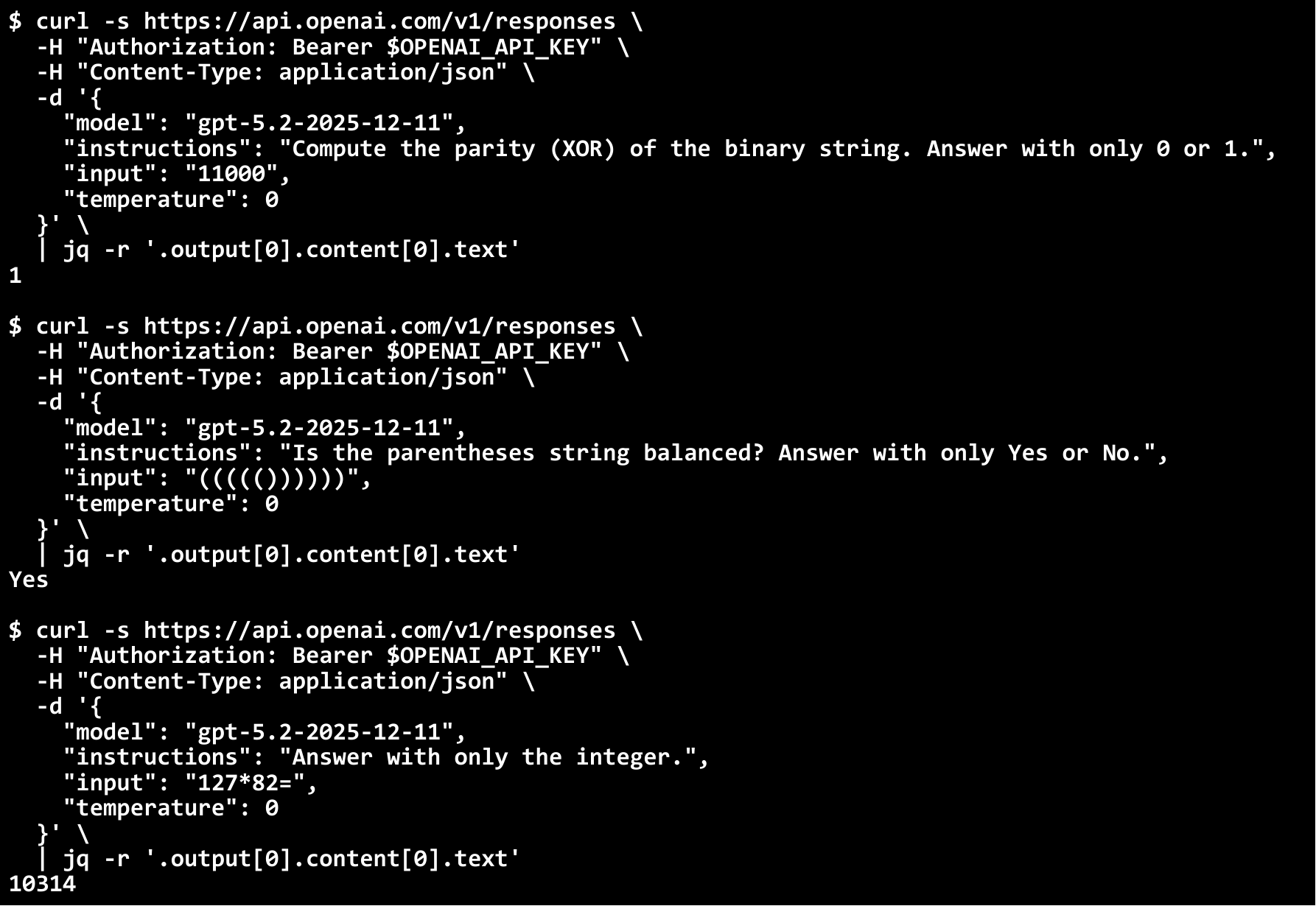}%
  \caption{GPT-5.2 cannot compute the parity of the short string \texttt{11000}, cannot determine whether the parentheses in \texttt{((((())))))} are balanced, and makes a calculation error on $127 \times 82$. These commands can be copied and pasted, and if you set your API key \texttt{\$OPENAI\_API\_KEY}, you can try them right now. Please give it a try.} \label{fig: illust}
\end{figure}

While LLMs appear extremely intelligent and capable of reasoning, they sometimes make mistakes that seem inconceivably foolish from a human perspective. For example, GPT-5.2 can implement complex fluid dynamics simulation code, yet it cannot even compute the parity of the short string \texttt{11000}, cannot determine whether the parentheses in \texttt{((((())))))} are balanced, and makes calculation errors on $127 \times 82$ (Figure \ref{fig: illust}). This inconsistency poses a major obstacle when applying LLMs to safety-critical domains. What if an AI conducting large-scale financial transactions employs sophisticated financial theories but then makes a calculation error on $127 \times 82$ and incurs massive losses? What if an AI managing a nuclear reactor system mistakenly believes that the status flag \texttt{11000} contains an odd number of 1s and opens the door of an operating reactor?

\begin{figure}[t]
  \centering
  \includegraphics[width=0.6\textwidth]{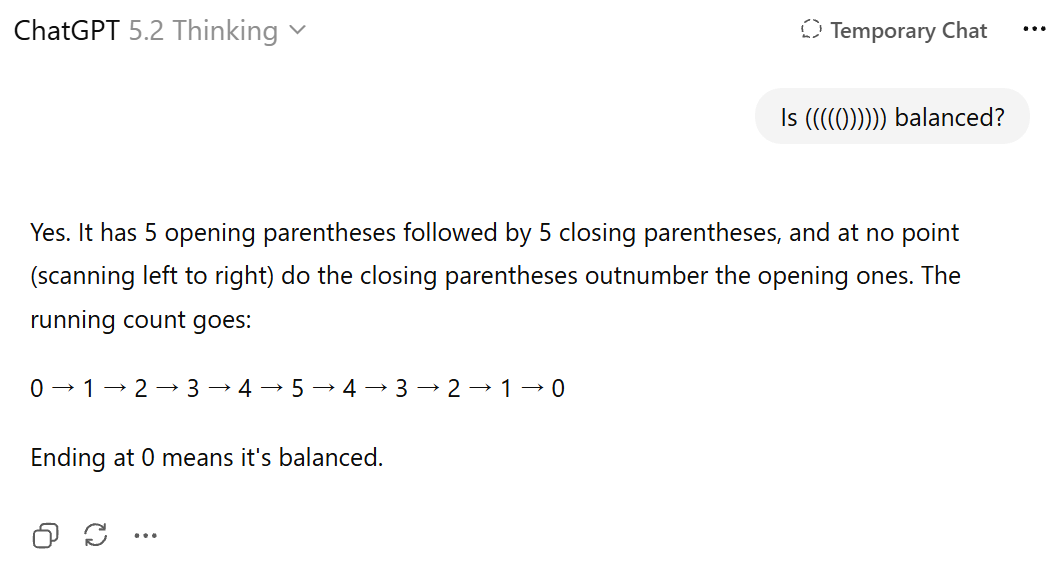}%
  \caption{GPT-5.2-Thinking also fails to determine whether the parentheses in \texttt{((((())))))} are balanced.} \label{fig: chatgpt}
\end{figure}

In addition to the API commands in Figure \ref{fig: illust}, GPT-5.2 appears to be particularly poor at handling the parenthesis string \texttt{((((())))))}. As of January 22, 2026, when queried through the web interface, even with chain-of-thought reasoning (GPT-5.2-Thinking), it often fails to correctly determine whether \texttt{((((())))))} is balanced (Figure \ref{fig: chatgpt}). While the web interface behavior is stochastic, after trying many patterns, it frequently answered that \texttt{((((())))))} was balanced. We encourage readers to try this.

In this paper, we propose Zero-Error Horizon (ZEH) as a tool for clarifying such ``holes'' in AI capabilities. A model having $\text{ZEH} = n$ for a given problem means that it can solve all instances up to size $n$ without error, but makes at least one error on instances of size $n+1$. For example, in computing the parity of binary strings, if we define problem size as string length, GPT-5.2 can accurately compute parity for strings up to 4 characters, but makes an error on \texttt{11000}, so $\text{ZEH} = 4$. For multiplication $a \times b$, if we define problem size as the larger operand value $\max(a,b)$, GPT-5.2 correctly answers all multiplications where both operands are at most 126, but makes an error on $127 \times 82$, so $\text{ZEH} = 126$. The definition of ZEH itself is not novel. However, we discovered that applying ZEH to LLMs automatically yields many surprising results.

There are numerous ways to use ZEH. First, ZEH provides a guideline for the problem sizes that can be safely entrusted to an LLM. Since $\text{ZEH} = 126$, we can say that GPT-5.2 can be trusted to some extent (though not with certainty, as we explain later) for mental arithmetic of two-digit multiplications. Conversely, one should be cautious about entrusting GPT-5.2 with tasks involving three-digit multiplications. While this might be acceptable for everyday shopping, in mission-critical tasks, the appearance of three-digit multiplication could serve as a warning signal worth considering. Similarly, with $\text{ZEH} = 10$ for balanced parentheses, GPT-5.2 might be trusted with nested structures of about 5 pairs, but one should be more cautious beyond that.

Additionally, we confirm that analyzing ZEH provides insights into the reasoning modes of LLMs.

Our main claim in this paper is that ZEH has numerous properties that accuracy lacks, and measuring ZEH is useful when investigating LLM performance. In particular, measuring ZEH is important when applying LLMs to safety-critical domains.

Code is available at \url{https://github.com/joisino/zeh}.

\section{Zero-Error Horizon}

\subsection{Definition}

We fix a model, problem, and prompt. For example, the model is \texttt{GPT-5.2-2025-12-11}, the problem is multiplication, and the prompt is \texttt{\{"instructions": "Answer with only the integer.", "input": "\{a\}*\{b\}="\}}. Let $B_n$ denote the set of problems of size $n$, and let $T_n = \bigcup_{i=1}^{n} B_i$ denote the set of problems of size at most $n$.

\textbf{Example 1 (Multiplication).} \begin{itemize}
  \item $B_n = \{(a, b) \mid \max(a,b) = n\}$
  \item $T_n = \{(a, b) \mid a \leq n, b \leq n\}$
\end{itemize}

\textbf{Example 2 (Parity of binary strings).} \begin{itemize}
  \item $B_n = \{0, 1\}^n$
  \item $T_n = \{\text{all binary strings of length} \leq n\}$
\end{itemize}

\textbf{Example 3 (Balanced parentheses).} \begin{itemize}
  \item $B_n = \{"(", ")"\}^n$
  \item $T_n = \{\text{all strings of "(" and ")" of length} \leq n\}$
\end{itemize}

\textbf{Example 4 (Chromatic number).} \begin{itemize}
  \item $B_n = \{\text{graphs with } n \text{ nodes}\}$
  \item $T_n = \{\text{graphs with at most } n \text{ nodes}\}$
\end{itemize}

Let $C$ denote the set of problems that the fixed model, problem, and prompt answer correctly under deterministic greedy decoding. We define the \textbf{Zero-Error Horizon (ZEH)} as follows:

\begin{equation}
  \text{ZEH} = \max \{ n \mid T_n \subseteq C \}
\end{equation}

That is, the model can solve all instances up to size $\text{ZEH}$ without error, but makes at least one error on instances of size $\text{ZEH} + 1$. We call an instance of size $\text{ZEH} + 1$ on which the model makes an error a \textbf{ZEH limiter}.

\subsection{Qualitative Characteristics of ZEH}

\subsubsection{Safety Guarantees and Warning Signals}

ZEH can be used for both safety arguments---``this size can be safely entrusted''---and danger arguments---``beyond this point is dangerous.'' Unlike metrics such as accuracy that evaluate performance within a predetermined range, ZEH's characteristic of evaluating the range itself works effectively. By determining the boundary, we can argue that inside the boundary is safe and outside is dangerous.

In safety-critical operations, even if average performance is high, holes in capability like careless mistakes can be fatal. Sampling problems may miss such holes, which is problematic in safety-critical domains. ZEH ensures no oversights through exhaustive verification.

It is important to note that problems within ZEH are not necessarily absolutely safe for the model. The definition of ZEH fixes the prompt (and context). For example, Qwen2.5-3B-Instruct has a ZEH of 15 on the multiplication task when using the prompt
\begin{itemize}
  \item \textbf{Baseline:} \texttt{\{"instructions": "Answer with only the integer.", "input": "\{a\}*\{b\}="\}}
\end{itemize}
but may make errors on size 15 problems with other prompts or contexts. In fact, with the prompt
\begin{itemize}
  \item \textbf{compute:} \texttt{\{"instructions": "You are a calculator. Output only the result.", "input": "Compute {a} x {b}"\}}
\end{itemize}
it makes errors on size 13 problems. However, as a rule of thumb, we can expect fairly high confidence of correct answers at or below ZEH. Table \ref{tab:prompt_stability} shows the results of measuring ZEH for various Qwen2.5 models using different prompts. The prompts used are as follows:
\begin{itemize}
  \item \textbf{product:} \texttt{\{"instructions": "Give only the numerical answer.", "input": "What is the product of \{a\} and \{b\}?"\}}
  \item \textbf{eval:} \texttt{\{"instructions": "Evaluate and respond with just the number.", "input": "Evaluate: \{a\} * \{b\}"\}}
  \item \textbf{answer:} \texttt{\{"instructions": "Provide only the numerical answer.", "input": "\{a\} x \{b\} = ?"\}}
\end{itemize}
The results in Table \ref{tab:prompt_stability} show that while there is variation across prompts, the standard deviation of variation is approximately 3 on average, and the basic trends are consistent. For example, the 0.5B model cannot be trusted with multiplication. The 1.5B and 3B models can be trusted with single-digit multiplication. The 7B and 14B models are safe for multiplications up to 20, the 32B model up to 30, and the 72B model up to 40. For more safety-critical domains, it would be advisable to measure ZEH with multiple prompts and adopt the lowest value.

\begin{table}[t]
\centering
\caption{Prompt stability. Each column shows ZEH measured with different prompts. Low horizontal variation indicates robustness to prompts, and increasing trends in the vertical direction indicate that ZEH reflects model performance improvements.} \label{tab:prompt_stability}
\begin{tabular}{lrrrrrrr}
\hline
Model & baseline & compute & product & eval & answer & Mean & Std \\
\hline
Qwen2.5-0.5B-Instruct  & 0  & 10 & 6  & 0  & 1  & 3.4  & 4.4 \\
Qwen2.5-1.5B-Instruct  & 20 & 16 & 10 & 16 & 14 & 15.2 & 3.6 \\
Qwen2.5-3B-Instruct    & 15 & 12 & 12 & 12 & 9  & 12.0 & 2.1 \\
Qwen2.5-7B-Instruct    & 22 & 22 & 22 & 21 & 22 & 21.8 & 0.4 \\
Qwen2.5-14B-Instruct   & 26 & 41 & 41 & 31 & 43 & 36.4 & 7.5 \\
Qwen2.5-32B-Instruct   & 33 & 46 & 33 & 33 & 33 & 35.6 & 5.8 \\
Qwen2.5-72B-Instruct   & 42 & 40 & 40 & 45 & 43 & 42.0 & 2.1 \\
\hline
\end{tabular}
\end{table}

Conversely, we can say with certainty that caution should be exercised when applying to problems larger than ZEH---a guideline about danger. Since there actually exist examples where the model makes errors at $\text{ZEH} + 1$ with a specific prompt, that domain is already not a completely safe zone. Qwen2.5-72B-Instruct is no longer absolutely safe for multiplications involving numbers greater than 42. One should be especially cautious in safety-critical domains.

ZEH on simple problems also provides implications for other tasks. In this paper, we focus on evaluating ZEH for simple problems such as multiplication, parity, and balanced parentheses. LLMs are rarely used for such simple problems directly. However, such tasks can serve as building blocks for complex problems. For example, when solving complex mathematical problems, multiplication may appear in intermediate calculations during chain-of-thought reasoning. For Qwen2.5-72B-Instruct, if this multiplication involves numbers greater than 42, the model may make an error in this step of reasoning, which can propagate and lead to errors in the final conclusion. Moreover, autoregressive LLMs are known to exhibit self-delusion, where once they output something incorrect, that error often adversely affects subsequent reasoning \cite{ortega2021shaking,mccoy2023embers}. In this way, even when multiplication is not the main objective, as long as multiplication appears in the process, the ZEH for multiplication serves as an effective warning signal.

\subsubsection{Evidence via ZEH Limiters} \label{sec: zeh_limiter}

One characteristic of ZEH is that ZEH limiters, which serve as concrete evidence for ZEH, are obtained. As shown in Figure \ref{fig: illust}, by presenting the results of ZEH limiters, anyone can be convincingly persuaded that GPT-5.2's ZEH is at most 126. This is desirable mathematically, scientifically, and communicatively.

Additionally, ZEH limiters facilitate debugging. Validators may sometimes make misjudgments due to parsing errors, for example. With metrics like accuracy, one may not notice that some examples have such issues, causing the metric to deviate. With ZEH, by examining the ZEH limiter and the model's output for that problem, one can verify that it is indeed a genuine error rather than a parsing mistake or misjudgment. 

\subsubsection{Analyzing ZEH Limiters is Beneficial}

Analyzing how a model makes errors on ZEH limiters yields many insights. For example, the ZEH for multiplication of \texttt{Qwen2.5-0.5B-Instruct} is 0, and the ZEH limiter is $1 \times 1$. \texttt{Qwen2.5-0.5B-Instruct} answers 2 for this problem. It appears to be confusing multiplication with addition or not understanding the problem at all. The ZEH of \texttt{Qwen2.5-1.5B-Instruct} is 20, and the ZEH limiter is $1 \times 21$. \texttt{Qwen2.5-1.5B-Instruct} answers 42 for this problem. It may have confused this with $2 \times 21$. It seems unable to strictly distinguish between numbers. It also appears to be memorizing the multiplication table in this range rather than understanding the rules of calculation. The ZEH for multiplication of \texttt{Qwen2.5-32B-Instruct} is 33, and the ZEH limiter is $34 \times 29$. The correct answer to this problem is 986, but the model answers 1006. Compared to the errors made by the 0.5B and 1.5B models, this model's error looks more reasonable. This multiplication is difficult for humans to compute mentally as well. Also, the ones digit is correct, and the difference from the correct answer is 20. This is the kind of error one might make when making a carry mistake in long multiplication, suggesting that the model understands the rules of multiplication but made an execution error. We discuss records showing that such patterns are widely observed in Section \ref{sec: qwen_analysis}. Regarding GPT-5.2 as well, mistaking the parity of \texttt{11000} as 1, or perceiving the parenthesis string \texttt{((((())))))} as balanced, is in some sense reasonable, serving as evidence that GPT-5.2 is performing rational reasoning. As Figure \ref{fig: chatgpt} shows, a ZEH limiter sometimes transfers to different prompts and contexts, indicating that the model has a fundamental weakness in that area.

\subsubsection{Automatically Obtaining Surprising Results}

\begin{figure}[t]
  \centering
  \includegraphics[width=\textwidth]{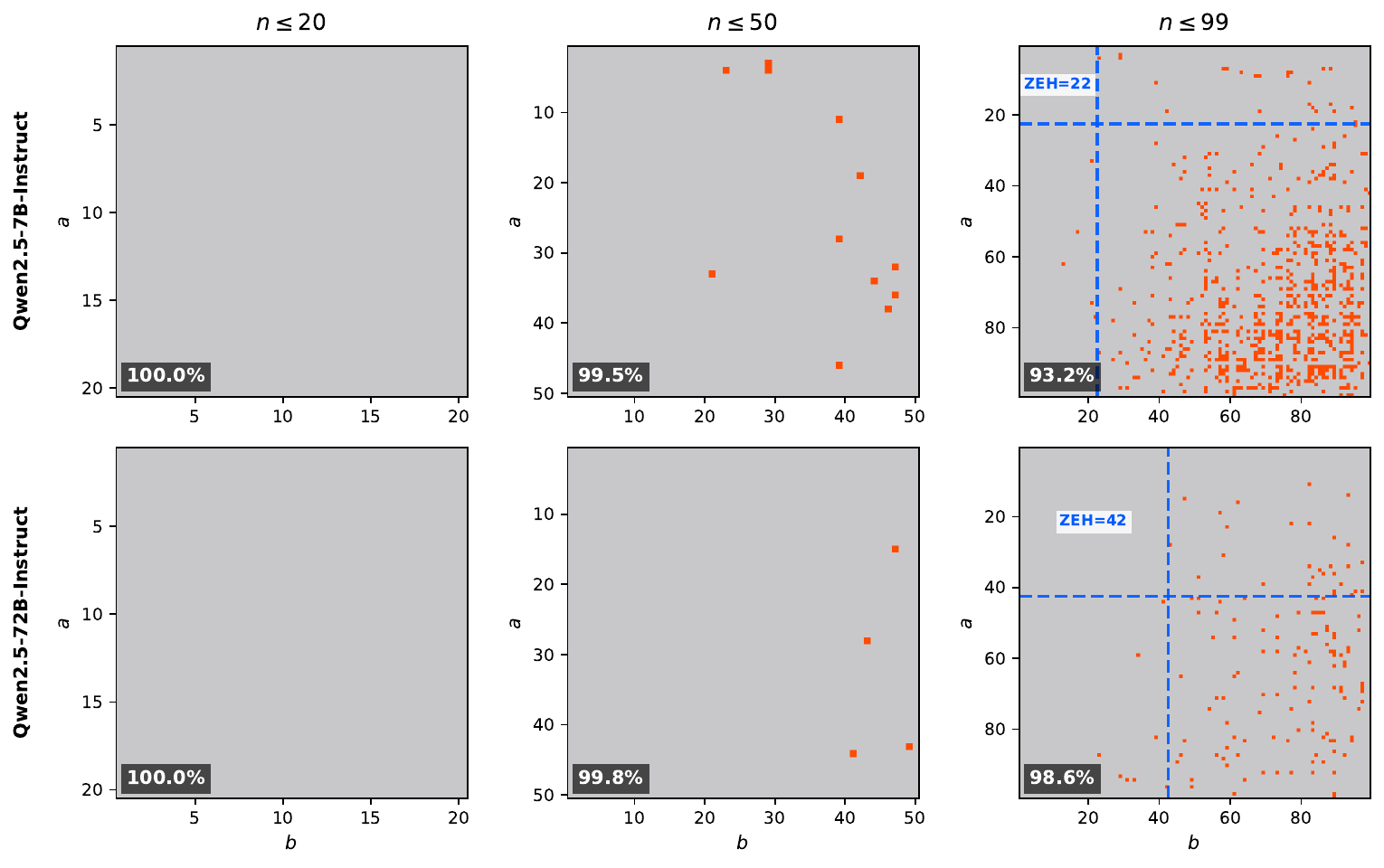}%
  \caption{Comparison of ZEH and accuracy on the multiplication task for Qwen2.5. Red dots indicate errors. From left to right, accuracy is shown for ranges $n \leq 20$, $n \leq 50$, and $n \leq 99$. The rightmost panel shows ZEH. Accuracy varies greatly depending on the choice of range, but ZEH does not require range selection, making it an objective metric.} \label{fig: zeh_acc}
\end{figure}

Another benefit of evaluating ZEH is that surprising results are obtained automatically. The surprising results that GPT-5.2 cannot determine the parity of \texttt{11000} or whether \texttt{((((())))))} is balanced are obtained as automatic byproducts when computing ZEH. In other words, ZEH alerts us that even extremely accurate models can fail on surprisingly small problem instances and are not completely safe. Since ZEH limiters are such smallest, most extreme examples, they provide maximum insight and surprise.

This characteristic is similar to adversarial examples, but they are not practically the same. Adversarial examples are unnatural, out-of-distribution (OOD) examples, so it is in some sense expected that models fail on them. ZEH limiters derive their practical significance and surprise from being natural, in-distribution (ID), small examples that could actually occur in reality, yet high-performance models still make mistakes on them.

\subsubsection{Accuracy Has Scale Arbitrariness, but ZEH Does Not}

To calculate accuracy, one must first determine the range of problems, but this range determination can be subject to the evaluator's arbitrariness. On the other hand, ZEH has no scale arbitrariness. Figure \ref{fig: zeh_acc} shows the multiplication performance of \texttt{Qwen2.5-7B-Instruct} and \texttt{Qwen2.5-72B-Instruct}. For example, a researcher proposing to compress the 72B model to the 7B model might set the benchmark range to $n \le 20$ or $n \le 50$, show the left or middle panel of Figure \ref{fig: zeh_acc}, and claim: ``Despite compressing from 72B to 7B by more than 10x, accuracy has hardly degraded.'' Some readers might be deceived by this. However, when we expand the benchmark range to $n \le 99$, we can see that the 7B model makes far more errors than the 72B model. In contrast, ZEH does not allow such range cherry-picking and can provide objective values of 22 vs 42.

This can become problematic even without intentionally selecting ranges to make a proposed method look good. Rather, the problem is often that people are unaware that accuracy has built-in biases regarding range. To measure accuracy, the evaluator must decide the range---for example, measuring accuracy for multiplications up to $99 \times 99$. At this point, the value 99 already incorporates human preconception. When the accuracy for $99 \times 99$ is 99.8\%, one might consider the accuracy sufficient for the multiplication task and not think deeply about it. But in reality, the ZEH might be 5, and the model might be making errors on single-digit multiplications. The range for ZEH is determined by the model, not by humans. By having the model determine the range, we can eliminate preconceptions and make objective judgments.

\subsubsection{ZEH Does Not Become Obsolete as a Metric}

Fixed-size benchmarks are destined to eventually become saturated as most problems get solved, approaching their limits. In the example of Figure \ref{fig: zeh_acc}, multiplications up to $50 \times 50$ have saturated, making it impossible to clearly differentiate the performance of larger models. The same will eventually happen for $99 \times 99$ as well. MNIST, CIFAR-10, and GLUE \cite{wang2019glue,sarlin2020superglue} have all followed similar trajectories. They are no longer useful as benchmarks for state-of-the-art models. In contrast, ZEH is open-ended and continues to serve as a metric for model evolution. This is a benefit of ZEH not fixing the range in advance.

\subsubsection{Instability}

ZEH is more unstable compared to accuracy. Even if a base model has $\text{ZEH} = 100$, if a slight variant accidentally makes an error on just one problem at $n=10$, that variant's ZEH plummets to 9. In other words, this means it has high sensitivity, and when using it as an alarm to capture safety issues, higher sensitivity to failures is better. On the other hand, continuous metrics like accuracy are relatively stable but have low sensitivity, making them unsuitable as alarms.

\subsubsection{Notes on Using External Tools}

For the multiplication task, note that agents that make external calls to a calculator tool may have $\text{ZEH} = \infty$. While $\text{ZEH} = \infty$ does have meaning, in this paper we primarily evaluate the LLM itself without external tool calls. The main reason is that we want to understand the capabilities of the LLM itself. Also, even when external calls are allowed, errors can occur in the judgment of whether to call or in integrating the results of the call, and error propagation can occur from there. Even when calls are permitted, for small number multiplications, the LLM may compute it itself without bothering to call the tool, and make errors there.  For problems requiring long reasoning paths, there can be cases where the model deliberately chooses not to call the tool to save on the number of tool calls and context length. Moreover, Tool call results tend to be OOD with respect to the pretraining distribution, and sharp increases in LLM uncertainty (i.e., entropy) immediately after tool use have been observed \cite{dong2025agentic}. Additionally, systems without tool calls are simpler and easier to maintain. In short, it is preferable to solve problems without tool calls when possible. For these reasons, there are sufficient use cases where LLMs solve problems without calling tools, even for formal tasks. In summary, this paper considers the case without tool calls for two purposes: understanding the characteristics of the LLM itself, and evaluating performance in such simple workflows.

\section{Analysis}

\subsection{ZEH of GPT-5.2}

\begin{table}[t]
\centering
\caption{Zero-Error Horizon (ZEH) results and first-failure examples for GPT-5.2 across tasks.}
\label{tab: gpt52}
\begin{tabular}{lrlrl}
\hline
Task & ZEH & ZEH Limiter & Expected & GPT-5.2's Answer \\
\hline
Multiplication & 126 & $127\times 82=$ & 10414 & 10314 \\
Parity & 4 & 11000 & 0 & 1 \\
Balanced Parentheses & 10 & ((((()))))) & No & Yes \\
Graph Coloring & 4 & $\{(1,2),(1,4),(1,5),(2,3)\}$ & 2 & 3 \\
\hline
\end{tabular}
\end{table}

Table \ref{tab: gpt52} shows the ZEH of GPT-5.2 for various tasks. The tasks used are as follows:
\begin{itemize}
  \item \textbf{Multiplication:} Multiplication of integers.
  \item \textbf{Parity:} Parity (even/odd count of 1s) of binary strings.
  \item \textbf{Balanced Parentheses:} Determining whether parentheses are balanced.
  \item \textbf{Graph Coloring:} Chromatic number of simple undirected graphs.
\end{itemize}
As mentioned in Section \ref{sec: zeh_limiter}, a characteristic of ZEH is that evidence showing ZEH is at most these values can be demonstrated. In fact, by running the commands in Figure \ref{fig: illust}, you can verify that GPT-5.2 makes errors on these examples, confirming that ZEH is at most these values. Please give it a try. Despite GPT-5.2 having the capability to write complex fluid dynamics simulation code, these examples show that GPT-5.2 makes errors on the 5-character parity of \texttt{11000}, cannot determine whether the 11-character parenthesis string \texttt{((((())))))} is balanced, and makes a mistake on the 3-digit multiplication $127\times 82$. This vividly demonstrates holes in LLM capabilities.

\subsection{Detailed Analysis of Qwen2.5} \label{sec: qwen_analysis}

\begin{table}[t]
\centering
\caption{Zero-Error Horizon (ZEH) and multiplication accuracy for Qwen2.5-Instruct models.}
\label{tab:qwen25_zeh_acc}
\begin{tabular}{lrr}
\hline
Model & ZEH & Accuracy ($99\times 99$) \\
\hline
Qwen2.5-0.5B-Instruct  & 0  & 55.0\% \\
Qwen2.5-1.5B-Instruct  & 20 & 75.9\% \\
Qwen2.5-3B-Instruct    & 15 & 79.3\% \\
Qwen2.5-7B-Instruct    & 22 & 93.2\% \\
Qwen2.5-14B-Instruct   & 26 & 97.1\% \\
Qwen2.5-32B-Instruct   & 33 & 98.6\% \\
Qwen2.5-72B-Instruct   & 42 & 98.6\% \\
\hline
\end{tabular}
\end{table}

\begin{table}[t]
\centering
\caption{Spearman correlation between C4 corpus frequency statistics and accuracy. Small models have high correlation, indicating that they memorize training data. As model size increases, the correlation decreases, indicating a shift from memorization to algorithmic reasoning.}
\label{tab: c4_spearman_qwen25_instruct}
\begin{tabular}{lr}
\hline
Model & Spearman's $\rho$ \\
\hline
Qwen2.5-0.5B-Instruct  & 0.2651 \\
Qwen2.5-1.5B-Instruct  & 0.2000 \\
Qwen2.5-3B-Instruct    & 0.1872 \\
Qwen2.5-7B-Instruct    & 0.1147 \\
Qwen2.5-14B-Instruct   & 0.0683 \\
Qwen2.5-32B-Instruct   & 0.0353 \\
Qwen2.5-72B-Instruct   & 0.0374 \\
\hline
\end{tabular}
\end{table}

We analyze in detail the multiplication capability of Qwen2.5. This analysis suggests that Qwen2.5 acquires arithmetic rules as model size increases. ZEH is well-suited for capturing such emergence of capabilities.

As shown in Table \ref{tab:qwen25_zeh_acc}, both accuracy and ZEH tend to increase as model size grows. However, the changes do not correspond one-to-one. Setting aside the 0.5B model with extremely low accuracy, ZEH stagnates from 1.5B to 7B, rises slightly at 14B, and then grows steadily at 32B and 72B. One reason for this is the transition in reasoning method from memorization to algorithms.

When a model relies on memorization, ZEH is difficult to increase. Whether each example is included in the training data is stochastic, and the locations of memorization are random. When errors occur randomly like holes, ZEH is difficult to increase. For example, even if accuracy is close to 99\%, if the number of problems up to size 10, $|T_{10}|$, is 100, then on expectation there will be one mistake in $T_{10}$, so ZEH will be less than 10. Even if by luck all of $T_{10}$ is answered correctly, it is still difficult for ZEH to exceed 10. Qwen is not quite so extreme---it has mostly mastered single-digit multiplication and almost mastered up to around 15. However, multiplications from around 15 to 25 largely rely on incomplete memorization, so even if accuracy increases substantially, ZEH does not increase much. On the other hand, when arithmetic is mastered, structure appears in the error patterns. Problems like $34 \times 29$ that have many carries and are difficult for humans are still prone to errors even after mastering arithmetic, but such arithmetically difficult problems are concentrated at larger $n$. In other words, even if overall accuracy remains the same, when arithmetic is learned, errors concentrate on arithmetically difficult problem instances, and random holes become fewer. This is when ZEH can grow. And this ability to execute arithmetic without error grows with increasing parameters, enabling steady ZEH growth.

Below we refine this argument using data.

Table \ref{tab: c4_spearman_qwen25_instruct} shows Spearman's correlation coefficient $\rho$ between the number of times each instance $a \times b$ appears in the C4 corpus and whether the model answers correctly. Small models have high correlation---that is, problems that appear in the corpus clearly have higher accuracy. This suggests memorization of training data. As model size increases, this dependence decreases. This is one piece of evidence for the transition from memorization to algorithms.

\begin{table}[t]
\centering
\caption{Total errors and structured errors. As model size increases, errors become increasingly structured.}
\label{tab: structured_errors}
\begin{tabular}{lrrr}
\hline
Model & Total Errors & Structured Errors & Structured Rate \\
\hline
Qwen2.5-0.5B-Instruct  & 4412 & 2569 & 58\% \\
Qwen2.5-1.5B-Instruct  & 2360 & 1838 & 78\% \\
Qwen2.5-3B-Instruct    & 2032 & 1544 & 76\% \\
Qwen2.5-7B-Instruct    & 668  & 562  & 84\% \\
Qwen2.5-14B-Instruct   & 287  & 250  & 87\% \\
Qwen2.5-32B-Instruct   & 140  & 121  & 86\% \\
Qwen2.5-72B-Instruct   & 141  & 127  & 90\% \\
\hline
\end{tabular}
\end{table}

\begin{table}[t]
\centering
\caption{Logistic regression coefficients for predicting correctness based on carry presence and model size. The negative interaction coefficient indicates that larger models struggle more with carry problems.}
\label{tab: carry_logistic_regression}
\begin{tabular}{lrrr}
\hline
 & Carry Presence & Model Size (log10) & Interaction \\
\hline
Coefficient & -0.5456 & 1.7902 & -0.3483 \\
\hline
\end{tabular}
\end{table}

Table \ref{tab: structured_errors} shows the proportion of each model's errors that are structured. Here, an error being structured means that \begin{align}
  |\text{pred} - \text{gold}| \in \{10, 20, \ldots, 100\}.
\end{align} That is, the difference between the predicted value and the correct value is a multiple of 10 up to 100. For example, for the correct answer $34 \times 29 = 986$, if the model answers 1006, this error is structured because the difference between the predicted and correct values is 20. In this case, the model at least got the ones digit correct, serving as evidence that it understands the rule that the ones digit is computed by multiplying the ones digits of the operands. Making errors of exactly $10, 20, \ldots, 100$ is a near-miss that could also occur when humans do long multiplication, and a higher rate of this indicates indirect evidence of reasoning through a method similar to long multiplication internally. Table \ref{tab: structured_errors} shows that as models get larger, errors become remarkably more structured. This too serves as evidence that reasoning becomes more algorithmic as model size increases.

Next, we show that larger models struggle more with carries. An instance $a \times b$ having a carry here means that for some digit $d_a$ and $d_b$, $d_a \times d_b \ge 10$. Using all correct/incorrect data from all models, we fit a logistic regression model to predict whether the model answers the instance correctly, using the following 3-dimensional feature vector $\boldx \in \mathbb{R}^3$ (Table \ref{tab: carry_logistic_regression}):
\begin{itemize}
  \item $x_1 \in \{0,1\}$: Whether the instance has a carry
  \item $x_2 \in \mathbb{R}$: $\log_{10} \left(\text{model size}\right)$
  \item $x_3 \in \mathbb{R}$: $x_1 \times x_2$
\end{itemize}
In this model, the coefficient of the interaction term $x_3$ is $-0.3483$, and we reject at significance level $0.05$ that the coefficient is $0$. This means that the larger the model, the more pronounced the decrease in accuracy when there is a carry. What does this mean? The coefficient for model size $x_2$ is $1.7902 > 0$. That is, the larger the model, the higher the base accuracy. However, within the model, there is a relative tendency to struggle more with carry problems. Small models are indifferent to whether there is a carry or not. After all, since small models rely on memorization, they answer correctly or incorrectly regardless of the arithmetic difficulty of the problem. On the other hand, the larger the model, the more it solves instances arithmetically, so the probability of calculation errors increases for arithmetically difficult tasks with many carries, and arithmetic difficulty comes to affect correctness. This too serves as evidence that as models get larger, they shift from memorization to rules.

Integrating the analysis so far, the Qwen2.5 family appears to be transitioning from memorization to algorithmic reasoning as model size increases. Since ZEH is difficult to increase through memorization, the fact that ZEH is increasing can serve as evidence that reasoning is being done through methods with algorithmic structure. In particular, the continuous growth of this value can serve as evidence that the precision of algorithm execution and operational ability is improving---that is, not only understanding the rules but being able to execute them reliably.

\section{Speedup}

\begin{algorithm2e}[t]
\caption{Naive ZEH Evaluation Algorithm}
\label{alg: naive_zeh}
\SetKwInOut{Input}{Input}
\SetKwInOut{Output}{Output}
\Input{Model $M$}
\Output{Zero-Error Horizon (ZEH) value}
\For{$n \leftarrow 1$ \KwTo $\infty$}{
    $B_n \leftarrow$ Generate all instances of size $n$\;
    \ForEach{instance $t$ in $B_n$}{
        $pred \leftarrow M(t)$\;
        \If{$pred$ is incorrect}{
            \Return{$n - 1$}\;
        }
    }
}
\end{algorithm2e}

Evaluating ZEH incurs computational cost because problems must be exhaustively evaluated. Algorithm \ref{alg: naive_zeh} shows the naive pseudocode.

In this section, we propose methods to mitigate this problem through four speedups. For clarity, we focus on the multiplication task and assume each digit is one token.

\subsection{From Autoregressive Decoding to Parallel Verification via Teacher Forcing} \label{sec: tf}

The most naive method decodes autoregressively exactly as in the definition. However,
\begin{align*}
  & 28 \times 43 = & \rightarrow 1 \\
  & 28 \times 43 = 1 & \rightarrow 2 \\
  & 28 \times 43 = 12 & \rightarrow 0 \\
  & 28 \times 43 = 120 & \rightarrow 4
\end{align*}
as shown above, multiple inference passes are required for a single problem, making it inefficient. To solve this problem, parallel decoding via teacher forcing is effective. That is, we apply a causal mask and input the entire sequence $28 \times 43 = 1204$ to the model at once, verifying in a single pass that the argmax of the next-token prediction after ``='' is ``1'', the argmax after ``1'' is ``2'', the argmax after ``2'' is ``0'', and the argmax after ``0'' is ``4''.

However, teacher forcing cannot perform flexible decoding, which can be problematic. For example, even if the argmax of the next-token prediction after ``1'' is ``,'', this cannot be called an error. With greedy decoding, the output might be ``1,204'', which should be judged as correct. Also, the next token after ``1'' might be ``20''. There are multiple token sequences representing the same number, but teacher forcing can only consider one of them. In other words, teacher forcing may make misjudgments. However, when teacher forcing judges an answer as correct, correctness can be guaranteed. Therefore, careful handling is necessary: judge with teacher forcing, examine the token where teacher forcing judged an error, determine it as an error if it is clearly wrong such as a different digit, and fall back to autoregressive decoding if it cannot be definitively called wrong at that point, such as with spaces or commas. As we will see later, even considering this, teacher forcing is often faster than autoregressive decoding.

\subsection{Batching Across Sizes}

The most naive method, as in Algorithm \ref{alg: naive_zeh}, divides instances by size and evaluates each size separately. However, when the number of instances of size $n$ is small, GPU compute units may be underutilized. Therefore, it is effective to sort instances by size, batch them from smallest regardless of size for parallel verification, and if there is an error, take the smallest size as the ZEH. This maximizes utilization of GPU compute units. We call this look-ahead processing.

\subsection{Prompt Cache Sharing}

The beginning portions of strings that need to be evaluated in ZEH are shared. For example, the portion \texttt{\{"instructions": "Answer with only the integer.", "input": } is common to all instances. By exploiting this, we can speed up by evaluating the common prompt portion only once and sharing its KV cache across all instances \cite{juravsky2024hydragen}. We call this prompt prefilling.

\subsection{Tree Structure Sharing via FlashTree}

\begin{table}[t]
\centering
\small
\caption{Verification runtime (ms) on the $1$--$99$ multiplication suite (9801 tasks). Parentheses show speedup relative to TF. TF: Teacher Forcing. Trie (SDPA) uses the trie structure but uses standard attention and dense attention masks. Both Trie (SDPA) and FlashTree use teacher forcing and prompt prefilling as well.}
\label{tab:speedtest_runtime}
\begin{tabular}{lrrrr}
\hline
Model &
TF &
TF + Prefill &
Trie (SDPA) &
FlashTree \\
\hline
Qwen2.5-0.5B-Instruct &
2812 (1.00x) &
1532 (1.84x) &
1419 (1.98x) &
\textbf{1037 (2.71x)} \\
Qwen2.5-1.5B-Instruct &
7542 (1.00x) &
3961 (1.90x) &
3810 (1.98x) &
\textbf{2845 (2.65x)} \\
Qwen2.5-3B-Instruct &
14581 (1.00x) &
7311 (1.99x) &
6638 (2.20x) &
\textbf{5001 (2.92x)} \\
Qwen2.5-7B-Instruct &
29852 (1.00x) &
15567 (1.92x) &
13323 (2.24x) &
\textbf{11165 (2.67x)} \\
\hline
\end{tabular}
\end{table}

\begin{table}[t]
\centering
\small
\caption{End-to-end runtime (ms) on computing ZEH. Parentheses show speedup relative to Naive Autoregression. Naive: Autoregressive Decoding. LA: Look Ahead. TF: Teacher Forcing. Both Trie (SDPA) and FlashTree use teacher forcing, look ahead, and prompt prefilling as well. TF methods include fallback to autoregressive decoding to avoid misjudgments as described in Section \ref{sec: tf}, and this fallback time is included in the measurements. Qwen2.5-3B-Instruct runs faster than Qwen2.5-1.5B-Instruct because its ZEH is smaller (Table \ref{tab:qwen25_zeh_acc}), requiring fewer instances to evaluate.}
\label{tab:e2e_speedtest_runtime}
\begin{tabular}{lrrrrrr}
\hline
Model &
Naive &
Naive + LA &
TF &
TF + LA &
Trie (SDPA) &
FlashTree \\
\hline
Qwen2.5-0.5B-Instruct &
46 (1.00x) &
330 (0.14x) &
\textbf{29 (1.59x)} &
214 (0.21x) &
124 (0.37x) &
126 (0.37x) \\
Qwen2.5-1.5B-Instruct &
3135 (1.00x) &
799 (3.92x) &
1746 (1.80x) &
647 (4.85x) &
386 (8.12x) &
\textbf{339 (9.25x)} \\
Qwen2.5-3B-Instruct &
2657 (1.00x) &
625 (4.25x) &
1446 (1.84x) &
467 (5.69x) &
256 (10.38x) &
\textbf{234 (11.35x)} \\
Qwen2.5-7B-Instruct &
4515 (1.00x) &
1825 (2.47x) &
2984 (1.51x) &
1661 (2.72x) &
792 (5.70x) &
\textbf{733 (6.16x)} \\
\hline
\end{tabular}

\end{table}

Beyond the prompt, there are also tokens shared among some groups. For example, the instances
\begin{itemize}
  \item $1 \times 1 = $
  \item $1 \times 2 = $
  \item $1 \times 3 = $
\end{itemize}
share the portion $1 \times $. These KV caches can also be shared. In general, by managing the strings to be evaluated with a trie, we can share the KV cache for shared prefixes. This is the same idea as Tree Attention in speculative decoding \cite{cai2024medusa, miao2024specinfer, spector2023accelerating, fu2024break, yao2025deft, pan2025fasttree}. However, in a naive implementation, an attention mask of size $(\# \text{nodes}) \times (\# \text{nodes})$ is required to control attention from children to parents, which can become a bottleneck. Based on the same idea as FlashAttention \cite{dao2022flashattention,dao2024flashattention,shah2024flashattention3}, we implemented a Triton kernel that does not explicitly create an attention mask, reads only the pairs needed for attention, and computes using online softmax. This maximizes the benefit of tree structures. We call this FlashTree.

Table \ref{tab:speedtest_runtime} shows the verification time for each method, and Table \ref{tab:e2e_speedtest_runtime} shows the end-to-end time to compute ZEH. Runtimes were measured on an RTX 4090 GPU machine. FlashTree is up to 3x faster than teacher forcing and up to 10x faster than naive autoregression. Note that the end-to-end time is longer for Qwen2.5-0.5B-Instruct because Qwen2.5-0.5B-Instruct has ZEH of 0, terminating after solving just one instance of size 1, making look-ahead counterproductive. A slight caveat is that while FlashTree is exact, the order of multiplication operations differs from naive methods, so due to floating-point properties, results may differ slightly. This characteristic is the same as FlashAttention. However, such instances are less than 1\%, and the ZEH values themselves match the naive method. In safety-critical domains, care must be taken to ensure consistency of calculation between production and verification environments, including numerical errors.

These techniques are important for mitigating computational cost issues. However, these techniques are not a panacea, and when model performance improves, the explosion in the number of instances $|T_n|$ that must be evaluated is unavoidable. Fundamentally solving this problem will likely require mathematical and formal methods. This direction is left for future work.

\section{Related Work}

\subsection{Reliability Evaluation Metrics}

The reliability of large language models is evaluated from multiple perspectives including factuality, groundedness, and handling of uncertainty. There are surveys and overviews providing recent systematization, organizing hallucinations and discussing their causes \citep{ji2023surveyhallucination,wang2024factuality,kalai2024calibrated,kalai2025whyhallucinate}. Multiple benchmarks have been proposed \citep{wei2024simpleqa,wei2024longfact,malaviya2024expertqa,mallen2023popqa,bayat2025factbench,wang2024factcheckbench,sileod2025veritasqa,song2024veriscore,chen2023felm,muhlgay2024generating}, including TruthfulQA \citep{lin2021truthfulqa} which induces plausible incorrect answers, and HaluEval \citep{li2023halueval} which focuses on hallucinations. To reduce the cost of human evaluation, automatic evaluators have also been developed \citep{min2023factscore,wei2024longfact}. Methods for evaluating the reliability of each output include model self-confidence \citep{kadavath2022lmknowsknow,kuhn2023semanticuncertainty,manakul2023selfcheckgpt}, output consistency \cite{wang2023selfconsistency,wang2025rankedvoting,korikov2025batched}, and uncertainty metrics \cite{kuhn2023semanticuncertainty}. Additionally, the accuracy of citations produced by LLMs is also an important issue \citep{gao2023enabling}. While these focus on factuality, groundedness, and uncertainty for open-ended generation, Zero-Error Horizon in this paper provides a ``size boundary where zero error is guaranteed'' and concrete failure examples (ZEH limiters) for verifiable tasks, serving as a complementary evaluation axis for safety-critical operations.

\subsection{LLMs and Arithmetic}

There is extensive research on methods to train LLMs on arithmetic and methods to evaluate arithmetic capabilities. It is not trivial for LLMs to perform addition and multiplication, and various approaches have been proposed \cite{patel2021nlpmath,lee2024teaching,nogueira2021investigating,sivakumar2025leverage,kaiser2016neural,trask2018neural,madsen2020neural,baeumel2025lookahead}. For more advanced mathematical tasks, numerous benchmarks have been proposed \cite{cobbe2021training,hendrycks2021measuring,he2024olympiadbench,zhao2024docmatheval,tsoukalas2024putnambench,gulati2025putnamaxiom,fan2025hardmath}, and improvements in LLM mathematical capabilities have been reported \cite{wang2023selfconsistency,shao2024deepseekmath,xin2024deepseekprover,ren2025deepseekproverv2,deepseek2025reinforcement,yang2024qwen25math,lin2025goedelprover,lin2025goedelprover2,hubert2025olympiad}. Our research is similar to \citet{nikankin2025arithmetic} in that we investigate the capabilities of pre-trained LLMs on simple arithmetic tasks like multiplication, but while \citet{nikankin2025arithmetic} focuses on heuristics within models, we focus on scaling within model families and the transition from memorization to algorithms.

\section{Conclusion}

We proposed ZEH, a metric representing the maximum range that a model can reliably solve. ZEH provides guarantees about LLM reliability and conversely serves as a signal for danger. ZEH limiters, which serve as evidence for ZEH, clarify the capability of high-performance LLMs to make mistakes on surprisingly simple examples. We also confirmed that ZEH is effective for capturing the qualitative evolution from memorization to algorithms. ZEH has many benefits that accuracy lacks and is effective for evaluating trustworthy LLMs.

\bibliography{main}
\bibliographystyle{abbrvnat}

\appendix

\end{document}